\documentclass[conference]{IEEEtran}
\IEEEoverridecommandlockouts
\usepackage{cite}
\usepackage{amsmath,amssymb,amsfonts}
\usepackage{algorithmic}
\usepackage{tabularx}
\usepackage[table]{xcolor}
\usepackage{colortbl}
\usepackage{multirow}
\usepackage{graphicx}
\usepackage{textcomp}
\usepackage{xcolor}
\usepackage{hyperref}
\usepackage{comment}
\usepackage{multicol,lipsum,microtype}
\hypersetup{
  colorlinks=true,
  linkcolor=black,  
  citecolor=black,
  urlcolor=black     
}
\def\BibTeX{{\rm B\kern-.05em{\sc i\kern-.025em b}\kern-.08em
    T\kern-.1667em\lower.7ex\hbox{E}\kern-.125emX}}
\begin{document}

\title{Classification of Non-native Handwritten Characters Using Convolutional Neural Network \\
}

\author{\IEEEauthorblockN{F. A. Mamun\textsuperscript{1}, S. A. H. Chowdhury\textsuperscript{2}, H. Sarker\textsuperscript{4}}
\IEEEauthorblockA{\textit{Dept. of Electronics \& Telecommunication Engineering} \\
\textit{Rajshahi University of Engineering \& Technology}\\
Rajshahi, Bangladesh \\
fayedruet@gmail.com\textsuperscript{1}, arif.1968.ruet@gmail.com\textsuperscript{2}, \\ hasan.ruet.ete@gmail.com\textsuperscript{4}}
\and
\IEEEauthorblockN{J. E. Giti\textsuperscript{3}}
\IEEEauthorblockA{\textit{Dept. of Electrical \& Electronic Engineering} \\
\textit{Rajshahi University of Engineering \& Technology}\\
Rajshahi, Bnagladesh \\
jishan.e.giti@gmail.com\textsuperscript{3}}
}
\maketitle


\begin{abstract}
The use of convolutional neural networks (CNNs) has accelerated the progress of handwritten character classification/recognition. 
Handwritten character recognition (HCR) has found applications in various domains, such as traffic signal detection, language translation, and document information extraction.
However, the widespread use of existing HCR technology is yet to be seen as it does not provide reliable character recognition with outstanding accuracy. 
One of the reasons for unreliable HCR is that existing HCR methods do not take the handwriting styles of non-native writers into account.
Hence, further improvement is needed to ensure the reliability and extensive deployment of character recognition technologies for critical tasks. 
In this work, the classification of English characters 
written by non-native users is performed by proposing a custom-tailored CNN model. We train this CNN with a new dataset called the handwritten isolated English character (HIEC) dataset. This dataset consists of 16,496 images collected from 260 persons. 
This paper also includes an ablation study of our CNN by adjusting hyperparameters to identify the best model for the HIEC dataset. 
The proposed model with five convolutional layers and one hidden layer outperforms state-of-the-art models 
in terms of character recognition accuracy 
and achieves an accuracy of $\mathbf{97.04}$\%. Compared with the second-best model, the relative improvement of our model in terms of classification accuracy is $\mathbf{4.38}$\%. 
\end{abstract}

\section{Introduction}
 Handwriting is a unique form of communication, with each individual's writing style being distinct and personal. Handwritten character recognition (HCR) systems, both online and offline, have gained prominence for their ability to analyze human handwriting across languages \cite{saqib2022convolutional}. 
 In recent years, these systems have seen increasing use in various applications, including reading postal addresses, language translation, bank forms, 
 digital libraries, and keyword spotting. 
 Convolutional Neural Networks (CNNs) are the reason behind this which offer the possibility of achieving highly robust and very accurate handwritten character classification (HCC). 
 
Inspired by human brain function, CNNs excel at recognizing patterns in images \cite{fukushima1980neocognitron}. 
Some well-known CNN architectures include 
AlexNet \cite{AlexNet_NIPS2012_c399862d}, VGGNet \cite{Simonyan15_VGG}, GoogLeNet \cite{Szegedy_GoogleNet_2015}, 
ResNet \cite{He_ResNet:2016}, InceptionV3 \cite{Szegedy_InceptionV3:2016}, MobileNet \cite{Howard2017MobileNetsEC}, and DenseNet \cite{Huang_DenseNet_2017} vary in the number of layers they possess, reflecting differences in depth and complexity.
CNNs are effective because they autonomously learn relevant features from data. Comprising convolutional, pooling, and fully connected layers, CNNs process input data, adapting internal parameters through back-propagation during training. Rigorous evaluation using separate test data ensures model accuracy.

\begin{figure}[hp]
\centering
\includegraphics[width=0.85\columnwidth]{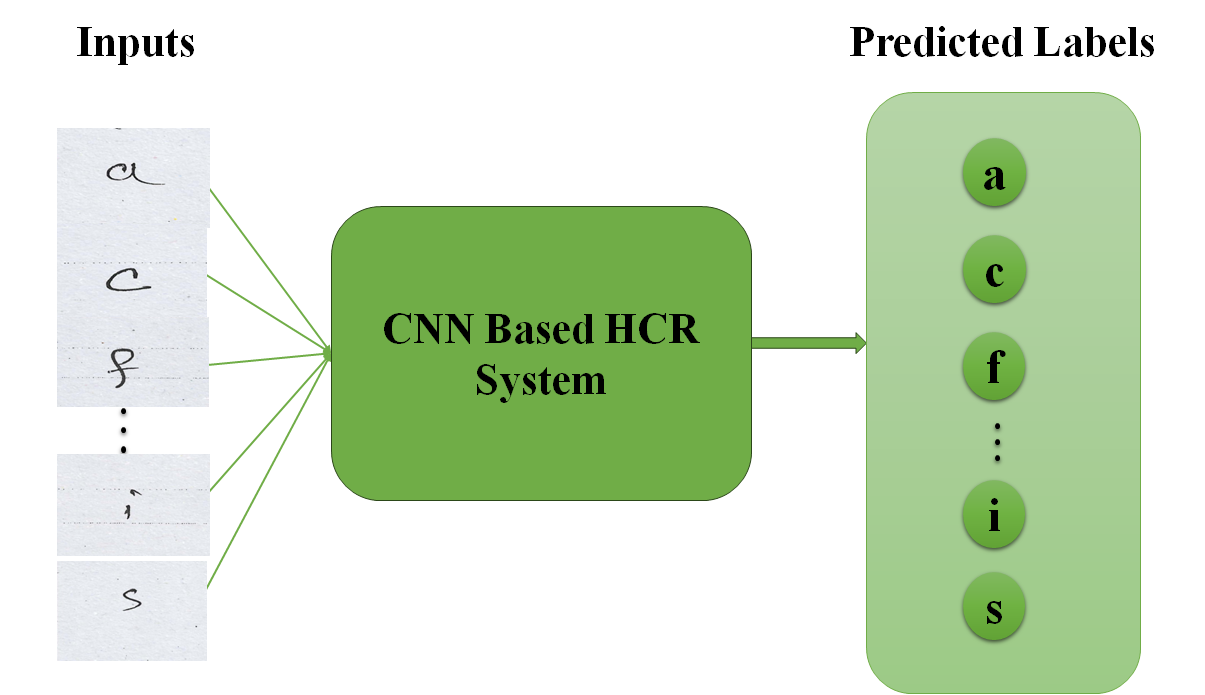}
\caption{An image of a handwritten character is being fed to a CNN-based handwritten character recognition (HCR) system. The system interacts with the input image and provides an output/predicted label. For accurate recognition, the predicted label needs to be similar to the ground truth label of the character being recognized. 
}
\vspace{-3mm}
\label{fig0}
\end{figure}

Fig. \ref{fig0} illustrates the process of a streamlined HCR system. It takes handwritten data (usually images) as input and seamlessly employs a CNN-based mechanism for precise recognition. The system's output provides accurately predicted labels, showcasing the efficacy of CNN's deep learning in deciphering diverse handwriting styles.

This paper aims to further explore the ability of CNNs for non-native HCC. To accomplish this, we use a new dataset called the handwritten isolated English character (HIEC) dataset
\footnote{\vspace{-0mm}The dataset is available at \url{https://tinyurl.com/5x9kjb68}.}
and design a benchmark CNN model\footnote{
The codes will be released at \url{https://tinyurl.com/2fydbdct}.} 
for this dataset. Our contributions are summarized as follows:
\begin{itemize}
    \item We collect an image dataset of non-native handwritten English lowercase letters called the HIEC dataset.
    \item We design a custom CNN to classify the characters of the HIEC dataset correctly.
    \item We conduct an ablation study of the proposed CNN to provide a better understanding of its learning process.
    \item We compare the classification performance of the proposed model with existing HCR models and state-of-the-art (SOTA) classification models.
\end{itemize}

Classification results demonstrate that our CNN-based HCR model achieves superior performance compared to SOTA models.  
The rest of the paper is organized as follows. Section II discusses existing related works and the research gap. In Section III, we describe the details of HIEC dataset collection. Section IV provides the methodology and the design procedure of the proposed CNN. Performance evaluation of our model along with a comparison with the existing models are shown in Section V. Finally, Section VI includes the concluding remarks as well as future works.

\section{Related Work}
Numerous new techniques have been introduced in research papers to classify texts.
These techniques can be broadly classified into two groups according to the types of text datasets used. The two different types of datasets are scene text and handwritten text datasets. We briefly review the CNN-based classification of scene and handwritten text. 

\textbf{Scene text.} Scene text recognition (STR) refers to identifying text content in natural scenes. Some scene text datasets are ICDAR 2003 \cite{ICDAR-2003}, street view text \cite{wang2011end}, SIW-13 \cite{SHI2016448}, and CVSI-2015 \cite{CVSI-2015}. These datasets contain scene images with mostly English text. An end-to-end solution using a multilayer CNN for STR has been investigated by Wang et al. \cite{wang2012end} where impressive results on benchmark datasets (street view text and ICDAR 2003) were reported. On the other hand, state-of-the-art results on SIW-13 and CVSI-2015 datasets were achieved by employing a discriminative CNN \cite{SHI2016448}. 

Improved scene text identification has resulted from combining a CNN and a recurrent neural network \cite{mei2016scene, Shi-2017-scene} which 
shows superior performance over the prior arts on benchmark datasets. 
Recent works on STR include permuted autoregressive sequence (PARSeq) \cite{bautista2022scene} and CLIP{$4$}STR \cite{zhao2023clip4str} models. 
CLIP{$4$}STR achieves state-of-the-art performance on benchmark datasets including COCO-Text \cite{veit2016coco} and IC19-Art \cite{chng2019icdar2019} whereas PARSeq ranked second. However, these STR models cannot be used for handwritten text classification without significant modification. 

\textbf{Handwritten text.} The focus of this paper is handwritten text classification. Some well-known handwritten text datasets are IAM \cite{Marti2002IAM}, MNIST \cite{deng2012mnist}, and EMNIST \cite{EMNIST}. Please note that all of these English handwritten datasets are collected from native writers. The IAM dataset contains images of lines of handwritten text whereas MNIST and EMNIST provide images of isolated text. MNIST is the digit dataset and extended MNIST (EMNIST) contains characters in addition to the digits of the MNIST dataset.

One of the recent works to achieve SOTA results on the MNIST dataset using a CNN without ensemble architecture has been carried out by Ahlawat et al. \cite{ahlawat2020improved}. On the contrary, EnsNet \cite{hirata2023ensemble} provides SOTA performance using the ensemble learning approach. 
Usually, ensemble learning requires more training parameters and training time.  

To decrease training parameters and epochs, capsule neural networks have been proposed \cite{sabour2017dynamic}. One of the limitations of the capsule neural network is that it requires dynamic routing between capsules. This limitation has been tackled using homogeneous vector capsules \cite{BYERLY2021545}. Capsule neural network obtains similar classification performance as the prior best CNN model on the MNIST dataset with fewer training parameters and epochs. These digit recognition models are not directly applicable to HCR and need to be fine-tuned with handwritten character images.  

Handwritten text recognition (HTR) of the IAM dataset requires an additional step of text segmentation before recognition as the dataset contains a sequence of text in a line. Two separate CNNs can be used to perform text segmentation and recognition. The pipeline of these two CNNs can be combined for simultaneous training of both networks to make the process end-to-end. Such an end-to-end HTR approach, namely TrOCR, has been proposed by Li et al. \cite{li2023trocr}. 

To reduce the complexity of the two-step process for HTR, it is preferable to be able to recognize the text without strictly applying segmentation to it. With this in mind, Singh et al. \cite{singh2021full} designed a CNN-based architecture for the recognition of full pages of handwritten or printed text which shows promising results without image segmentation. However, the presence of the separate segmentation step provides an opportunity for the recognition models to be equally applicable to isolated (MNIST and EMNIST) and sequential (IAM) datasets.


SOTA results on the unbalanced EMNIST dataset have been achieved by Saqib et al. \cite{saqib2022convolutional} using a custom CNN model. By splitting each layer into three splits, SpinalNet \cite{kabir2022spinalnet} provides a competitive performance compared to the current SOTA model on the EMNIST dataset called WaveMix \cite{jeevan2022wavemix}. Multi-level two-dimensional discrete wavelet transform is employed in WaveMix blocks which has several advantages including scale-invariance, shift-invariance, and sparseness of edges. 
Another work on the EMNIST dataset has been carried out by Mor et al. \cite{mor2019handwritten} where an application for Android is developed for HTR. The English HTR models mentioned so far are tested with the handwriting of native writers and the effect of diverse handwriting styles of non-native writers is yet to be seen.


Besides English handwritten text, studies have showcased the efficacy of deep CNNs for Devanagari \cite{jangid2018handwritten}, Bangla \cite{alif2017isolated}, Shui \cite{weng2020new} and
Gujarati \cite{joshi2018deep} character recognition.  
More promising CNN-based classification results on highly cursive Urdu ligatures and Arabic characters have been reported by Javed et al. \cite{javed2017classification} and Elleuch et al. \cite{elleuch2016new}, respectively. Furthermore, word-level script identification using CNN for $11$ different languages has been explored by Ukil et al. \cite{ukil2021deep}. In addition, transfer learning with CNN has been employed to recognize numerals of various languages \cite{tushar2018novel}.



In summary, CNN can successfully classify handwritten texts of different styles and languages. However, in most cases, these handwritten texts are written by native users of the language. The effect of non-native writers on existing HCR models is not yet considered. 
In this paper, we investigate the effect of non-native writers on HCR by collecting the HIEC dataset and designing a SOTA model for the dataset.

\section{Dataset Collection and Description}
In this section, we provide details about the collection process and characteristics of the HIEC dataset obtained from the students of RUET, a prestigious engineering University.

\subsection{Dataset Collection}
\begin{figure}[!hp]
\centering
\includegraphics[width=0.8\columnwidth]{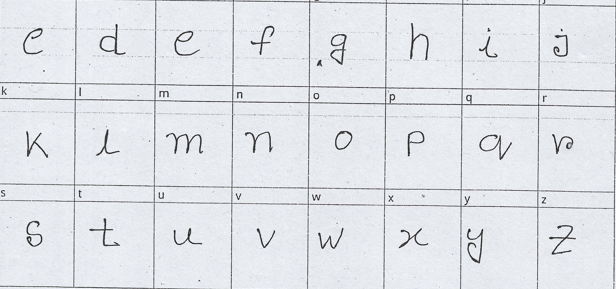}
\caption{Some samples of the HIEC dataset.}
\vspace{-3mm}
\label{fig4}
\end{figure}

The HIEC dataset was collected from 260 students and all of these individuals were non-native English writers. The students were provided with writing materials, such as pens or pencils, and asked to write down a set of English lowercase letters on paper. Each writer contributed their handwritten characters, resulting in a dataset of 6,760 samples. To capture the natural variability present in handwritten English characters, the students were instructed to write the characters in their natural handwriting style, without any specific guidelines or constraints. Fig. \ref{fig4} shows some samples of the collected dataset. 

\subsection{Dataset Description}
The collected dataset encompasses diverse images 
of English lowercase letters. The letters were written in various sizes, shapes, and styles, reflecting the diverse writing habits and preferences of non-native English writers. By encompassing a wide range of writing habits, 
the HIEC dataset 
provides a comprehensive representation of handwritten English characters encountered in practical applications.

This new dataset contains both challenging samples, where the characters might be written quickly or with slight distortions, as well as easier-to-read samples with clear and well-formed characters. Some students may have written certain characters faster than others, which could introduce variations in the shape and quality of those characters. This aspect of the dataset reflects real-world scenarios where handwriting can vary due to the speed at which individuals write. Thus, the collected dataset creates an opportunity to evaluate how accurately the model can recognize characters written hastily.
\section{Methodology}
\begin{figure}[!h]
\centering
\includegraphics[width=\columnwidth]{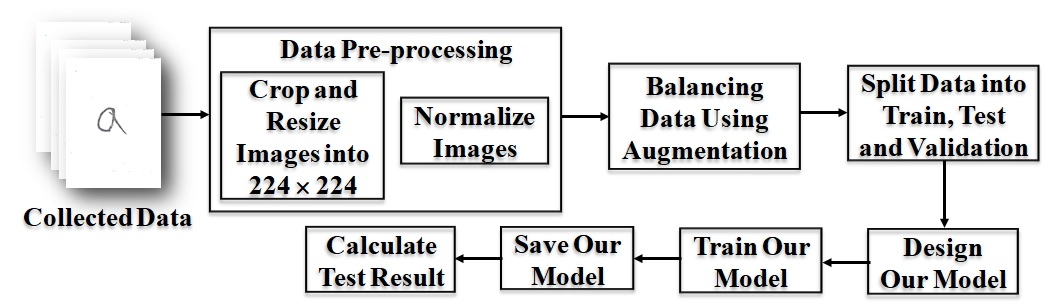}
\caption{The overall pipeline of this work begins with a pre-processing step where the collected images of handwritten characters are cropped, then resized to a standard size, and normalized. The next step is data augmentation to balance and increase the size of the HIEC dataset. After splitting the augmented dataset into training, validation, and test sets, we design and train our model using training and validation data. The best-trained model is then saved and used to obtain the classification result on the test set.
}
\vspace{-3mm}
\label{fig1}
\end{figure}

CNN has revolutionized the field of computer vision and has been widely used in various applications, including image classification, object detection, and, in our case, HCR. CNN excels at capturing local patterns and hierarchical representations in visual data, making them particularly suitable for tasks like HCR. 
The overall pipeline of the methodology for CNN-based HCR is displayed in Fig. \ref{fig1}.

\subsection{Data Preprocessing}
To ensure compatibility with deep learning models, we performed preprocessing steps on the collected dataset. First, we cropped the images to remove unnecessary background information, focusing solely on the handwritten characters. This step helps the model to concentrate on the characters' shape and structure. Subsequently, we resized the cropped images to a consistent size so that all images could be transformed into a uniform format for model training and inference. Then, normalization is done to keep all pixel values within the desired range. 
Normalization also facilitates better model performance by reducing the influence of varying pixel intensity ranges across different images.

\subsection{Data Augmentation}
\begin{figure}[!h]
\centering
\includegraphics[height=0.55\columnwidth,width=0.55\columnwidth]{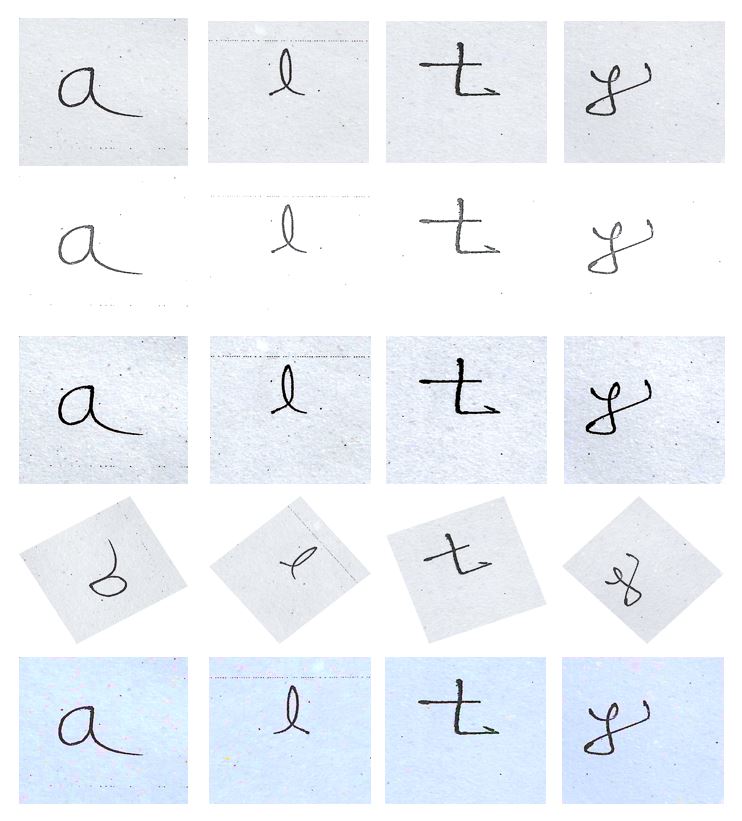}
\caption{Examples of images after applying various augmentation techniques. Each row represents images corresponding to different augmentation techniques. The top row shows the original data. The second, third, fourth, and fifth rows from the top show the augmented images after applying brightness adjustment, contrast adjustment, rotation, and sharpness adjustment, respectively. 
}
\label{fig2}
\end{figure}

Data augmentation is a widely used approach in machine learning, particularly when the dataset is limited in size. For our HIEC dataset, which contains only 6760 samples, data augmentation becomes vital to expand its volume. 
If we trained our model using a larger dataset, then our model would give better performance because suitable training of CNN claims big data \cite{rahman2020transfer}. Our augmented dataset has 16,496 samples. In this work, we have used four augmentation methods. Fig. \ref{fig2} shows some examples of augmented images corresponding to different augmentation.


\begin{figure*}[t]
\centering
\includegraphics[width=0.95\textwidth,height=5.8cm]{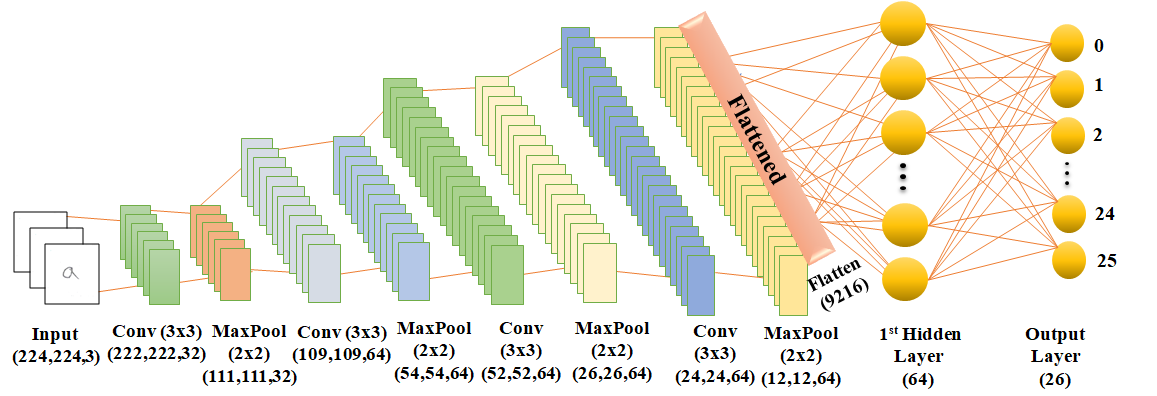}
\caption{In the proposed CNN model, the same kernel and pooling sizes are used for each conv and max-pooling layer, respectively. The size of the kernels for convolution and max-pooling are $3\times3$ and $2\times2$, respectively. 
The handwritten character images are fed to the first convolutional layer as input.
The output size of each layer is shown inside the bracket at the bottom of the illustration corresponding to that layer. For example, the output size of the second convolutional layer is $109\times109\times64$ whereas the output size after the last max-pooling layer is $12\times12\times64$. The output from the last max-pooling layer is flattened which yields the size of $9216\times1$. This output serves as the input to the subsequent dense layers. The first dense (hidden) layer contains $64$ neurons, whereas the last dense layer, serving as the output layer, has $26$ neurons equal to the number of classes in the dataset. The softmax activation function is chosen for the output layer, transforming the raw output scores into a probability distribution over the classes. 
}
\label{fig3}
\end{figure*}

\textbf{Brightness and Contrast.} To introduce variations in illumination conditions, we applied brightness and contrast adjustments. 
Thus, the model learns 
characters under different lighting conditions, making it more robust and adaptable.

\textbf{Rotation.} Rotation was applied to introduce variations in the orientation of the characters. 
In consequence, orientation-independent character recognition can be achieved which enables the model to generalize to various writing styles. 

\textbf{Sharpness.} The sharpness variation was employed to enhance the edge details of the characters. 
This augmentation helps the model capture intricate features and improve the recognition accuracy.


\subsection{Proposed Model}

The proposed CNN architecture is designed to recognize handwritten English characters efficiently and effectively. Our base model comprises four convolutional (conv) layers, each followed by a max-pooling layer. 
ReLU activation is applied after each conv layer and the hidden layer, introducing non-linearity and enhancing the model's ability to learn complex representations from higher-level features.

 The architecture of our base CNN model is shown in Fig. \ref{fig3}.
Throughout the model's architecture, the output size is effectively managed, 
while capturing intricate features crucial for accurate HCR. The progressive reduction in spatial dimensions during convolution and max-pooling operations allows the model to efficiently learn and generalize across the dataset, 
achieving high recognition accuracy.



\subsection{Performance Metrics}
The following performance metrics have been used to compare the classification
results in this study: accuracy, precision, recall, and F1-score \cite{goutte2005probabilistic}. Please note that these performance metrics reported throughout the paper are for the testing dataset unless mentioned otherwise.

\subsection{Implementation Details}
Our model is trained from scratch using Adam optimizer \cite{DBLP:journals/corr/KingmaB14} with a learning rate of $0.001$ and batch size of $64$. We train our model using cross-entropy loss on Kaggle, a data science competition platform. Approximately $80$\% ($13,168$ images) and $10$\% ($1,536$ images) of the augmented data are used for training and validation, respectively, whereas the remaining $10$\% ($1,792$ images) are used for testing. 

\section{Results \& Analysis}
\label{R}

This section contains an ablation study of the proposed CNN model and a comparison of our model with SOTA HCR models.
An ablation study provides insights into the significance of each layer's contribution to the overall network's effectiveness.


\subsection{Ablation Study}
An ablation study is conducted by changing the number of filters in conv layers, the number of conv layers, max-pooling layers, and hidden layers of our base model. The results of the ablation study are summarized in Table \ref{tab1} where the first and fourth rows from the top correspond to the base and best models, respectively. 

The accuracy of our base model is $94.64$\%. 
We found better accuracy with slightly different configurations of the base model.
Our best model has five conv and max-pooling layers, one hidden layer followed by a dropout layer to prevent overfitting, and achieves an accuracy of $97.04$\% which is highlighted in Table \ref{tab1}. 

\begin{table}[!h]
    \vspace{-5mm}
    \renewcommand{\arraystretch}{1.2} 
    \setlength{\tabcolsep}{0.07cm} 
    \caption{Comparison of Different CNN configurations in terms of testing accuracy, precision, recall, and F$1$-score. In the first and second left-most columns, the values outside the brackets are the number of conv layers and hidden layers, respectively. The values inside the brackets indicate the corresponding number of kernels and neurons in each consecutive conv and hidden layer, respectively. }
    \centering
    \label{tab1}
    \begin{tabular}{|c|c|c|c|c|c|c|}
        \hline 
        \begin{tabular}{@{}c@{}} No. of \\ conv layers\\  and kernels 
        \end{tabular} 
        & \begin{tabular}{@{}c@{}} No. of \\ hidden \\ layers \\ and \\ neurons \end{tabular} 
        & \begin{tabular}{@{}c@{}} Dropout \\ (\%) \end{tabular} & \begin{tabular}{@{}c@{}} Accuracy \\ (\%) \end{tabular} & \begin{tabular}{@{}c@{}} Precision \\ (\%) \end{tabular} & \begin{tabular}{@{}c@{}} Recall \\ (\%) \end{tabular} & \begin{tabular}{@{}c@{}} F1-Score \\ (\%) \end{tabular} \\
        \hline 
        \begin{tabular}{@{}c@{}} $4$ ($32$, $64$, \\ $64$, $64$) \end{tabular} 
        & \begin{tabular}{@{}c@{}} $1$ ($64$) \end{tabular} 
        & $0$ & $94.64$ & $93.92$ & $94.01$ & $93.96$ \\
        \hline 
        \begin{tabular}{@{}c@{}} $3$ ($32$, $64$, \\ $128$) \end{tabular} 
        & \begin{tabular}{@{}c@{}} $3$ ($512$, \\ $256$, $128$) \end{tabular} 
        & $50$ & $93.86$ & $93.52$ & $93.88$ & $93.69$ \\
        \hline 
        \begin{tabular}{@{}c@{}} $4$ ($32$, $64$, \\ $128$, $256$) \end{tabular} 
        & \begin{tabular}{@{}c@{}} $2$ ($512$, \\ $256$) \end{tabular} 
        & $50$ & $96.43$ & $96.52$ & $96.88$ & $96.69$ \\
        \hline 
        \begin{tabular}{@{}c@{}} $\mathbf{5}$ ($\mathbf{64}$, $\mathbf{64}$, \\ $\mathbf{128}$, $\mathbf{512}$, \\ $\mathbf{128}$) \end{tabular} 
        & \begin{tabular}{@{}c@{}} $\mathbf{1}$ ($\mathbf{256}$) \end{tabular} 
        & $\mathbf{50}$ & $\mathbf{97.04}$ & $\mathbf{97.02}$ & $\mathbf{97.02}$ & $\mathbf{97.02}$ \\
        \hline 
        \begin{tabular}{@{}c@{}} $6$ ($32$, $64$, \\ $64$, $64$, \\ $64$, $64$) \end{tabular} 
        & \begin{tabular}{@{}c@{}} $1$ ($64$) \end{tabular} 
        & $0$ & $95.92$ & $95.94$ & $95.88$ & $95.89$\\
        \hline
    \end{tabular}
\end{table}

The accuracy of $97.04$\% means that out of $1792$ test images of the HIEC dataset, $1739$ character images are recognized accurately by our best model.
In this case, the total true positive is $1747$ when the model correctly identifies a character from a particular class. For the remaining $53$ images, the model falsely or correctly identifies a character as being or as being not part of that particular class contributing to the total false positive or total false negative, respectively. 

\begin{figure}[!h]
\centering
\includegraphics[width=0.95\columnwidth]{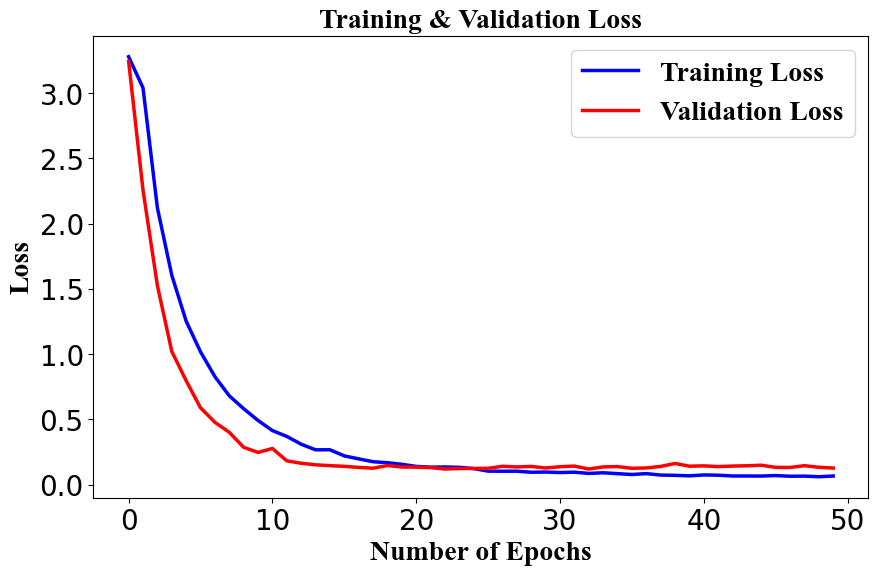}
\caption{Traning and validation loss vs. epoch curves of our best model. 
}
\vspace{-3mm}
\label{fig6}
\end{figure}

The training and validation loss curve of our best model is shown in Fig. \ref{fig6}. The validation loss follows the training loss as expected according to Fig. \ref{fig6}. It can also be seen from Fig. \ref{fig6} that, the losses stay almost the same after $30$ epochs. This suggests that the training process converges around $30$ epochs.

\subsection{Performance Comparison with Related Works}

We compare the performance of our model with several popular classification models \cite{He_ResNet:2016, Szegedy_InceptionV3:2016, Howard2017MobileNetsEC, Huang_DenseNet_2017}
as well as existing state-of-the-art HCR models \cite{mor2019handwritten, weng2020new, joshi2018deep}. 
The objective of the comparison is to assess the potential of our model in detecting handwritten characters of the HIEC dataset. Comparison results are provided in Table \ref{tab3}. 
Through this comparison, we found that our model outperforms the popular classification models as well as state-of-the-art HCR models in terms of accuracy, precision, recall, and F1-score. 

\begin{table}[!h]
\vspace{-5mm}
 \renewcommand{\arraystretch}{1.2} 
  \setlength{\tabcolsep}{0.1cm} 
    \caption{Comparison of our model with state-of-the-art classification models and handwritten character recognition models in terms of different performance metrics. The relative improvement of our model compared to the second-best model \cite{weng2020new} in terms of accuracy, precision, recall, and F1-score is $4.38$\%, $4.57$\%, $4.57$\%, and $4.57$\%, respectively.}

  \centering
  \begin{tabular}{|c|c|c|c|c|}
    \hline
    Model & Accuracy & Precision & Recall & F1-Score \\
    \hline
    ResNet50 \cite{He_ResNet:2016} & $91.02$\% & $91.01$\% & $91.03$\% & $91.02$\% \\
    \hline
    InceptionV3 \cite{Szegedy_InceptionV3:2016} & $84.20$\% & $84.15$\% & $84.15$\% & $84.15$\% \\
    \hline
    MobileNet \cite{Howard2017MobileNetsEC}& $81.69$\% & $81.69$\% & $81.69$\% & $81.69$\% \\
    \hline
    DenseNet121 \cite{Huang_DenseNet_2017} & $80.13$\% & $80.12$\% & $80.14$\% & $80.13$\% \\
    \hline
    Mor et al. \cite{mor2019handwritten} & $89.73$\% & $89.52$\% & $89.54$\% & $89.53$\% \\
    \hline 
    Weng and Xia \cite{weng2020new} & $92.97$\% & $92.78$\% & $92.78$\% & $92.78$\% \\
    \hline
    Joshi and Risodkar \cite{joshi2018deep} & $41.3$\% &$65$\% & $41$\% & $47$\%\\
    \hline
    \textbf{Ours} & $\mathbf{97.04}$\textbf{\%} &$\mathbf{97.02}$\textbf{\%} &$\mathbf{97.02}$\textbf{\%} &$\mathbf{97.02}$\textbf{\%}\\
    \hline
  \end{tabular}
  \label{tab3}
\end{table}

As existing HCR models are designed for native handwriting classification, they struggle to learn non-native handwritten characters of the HIEC dataset.
The simplicity and lightweight nature of our CNN have proven to be advantageous, especially considering that our dataset is not extensive. 
Our model's ability to capture relevant features from the limited dataset contributed significantly to its superior performance compared to the popular classification models. The relative improvement of our model compared to 
ResNet50, InceptionV3, MobileNet, and DenseNet121 models in terms of recognition accuracy is $6.61$\%, $15.23$\%, $18.77$\% and $21.08$\%, respectively.The superior performance of our custom CNN model over all others model can be attributed to its tailored architecture optimized for the specific characteristics of our image dataset, complemented by the efficiency derived from its lightweight design.

\section{Conclusion and Future Work}

In this paper, we attempt to tackle the problem of handwriting recognition of non-native users. To the best of our knowledge, this paper is the first research work to address this issue. We tackle the problem by demonstrating state-of-the-art classification performance using a proposed CNN model on a handwriting dataset collected from non-native individuals. The effectiveness of our model for non-native handwritten character classification has been reported through a comprehensive comparison. Our model's ability to handle diverse handwriting styles of non-native writers contributes to the advancement of character recognition technology for real-world applications. 

As part of our future research endeavors, we plan to extend this work to non-native handwritten multilingual character recognition.
In addition, our intention is to apply a similar framework to address the challenges posed by more intricate languages, including Korean, Chinese, Finnish, and Japanese. Furthermore, employing a semi-supervised learning algorithm to achieve state-of-the-art performance on the HIEC dataset is also our future goal.

\section*{Acknowledgement}
The authors would like to thank all the participating students of Rajshahi University of Engineering \& Technology (RUET), Bangladesh who contributed their handwriting to the collection of the HIEC dataset.

\bibliographystyle{IEEEtran}
\bibliography{conference_101719}

\end{document}